\documentclass[runningheads]{llncs}
\usepackage{graphicx}

\usepackage{tikz}
\usepackage{comment}
\usepackage{amsmath,amssymb} 
\usepackage{color}

\usepackage[accsupp]{axessibility}  

\usepackage{cite}
\usepackage{subcaption}

\usepackage{multirow}
\usepackage{booktabs,tabularx}
\usepackage{siunitx}
\newcolumntype{C}{>{\centering\arraybackslash}X}

\newcommand*{\rowstyle}[1]{
  \gdef\@rowstyle{#1}%
  \@rowstyle\ignorespaces%
}

\usepackage[colorlinks,linkcolor=black,citecolor=black,urlcolor=black]{hyperref}

\begin{document}
\pagestyle{headings}
\mainmatter
\def\ECCVSubNumber{7651}  

\title{SPE-Net: Boosting Point Cloud Analysis via Rotation Robustness Enhancement} 

\titlerunning{Selective Position Encoding Networks}
\author{Zhaofan Qiu$^\star$\and
 Yehao Li\thanks{Zhaofan Qiu and Yehao Li contributed equally to this work.} \and
 Yu Wang \and
 Yingwei Pan \and
 Ting Yao \and
 Tao Mei}
\authorrunning{Z. Qiu, Y. Li, Y Wang, Y. Pan, T. Yao, and T. Mei}
%
\institute{JD Explore Academy, Beijing, China \\
\email{\{zhaofanqiu, yehaoli.sysu, feather1014, panyw.ustc, tingyao.ustc\}@gmail.com; tmei@jd.com}}

\maketitle
\begin{abstract}
In this paper, we propose a novel deep architecture tailored for 3D point cloud applications, named as SPE-Net. The embedded ``Selective Position Encoding (SPE)'' procedure relies on an attention mechanism that can effectively attend to the underlying rotation condition of the input. Such encoded rotation condition then determines which part of the network parameters to be focused on, and is shown to efficiently help reduce the degree of freedom of the optimization during training. This mechanism henceforth can better leverage the rotation augmentations through reduced training difficulties, making SPE-Net robust against rotated data both during training and testing. The new findings in our paper also urge us to rethink the relationship between the extracted rotation information and the actual test accuracy. Intriguingly, we reveal evidences that by locally encoding the rotation information through SPE-Net, the rotation-invariant features are still of critical importance in benefiting the test samples without any actual global rotation. We empirically demonstrate the merits of the SPE-Net and the associated hypothesis on four benchmarks, showing evident improvements on both rotated and unrotated test data over SOTA methods. Source code is available at \href{https://github.com/ZhaofanQiu/SPE-Net}{https://github.com/ZhaofanQiu/SPE-Net}.
\end{abstract}

\section{Introduction}

Pioneering efforts on 3D cloud point analysis have paid much attention on dealing with rotated data. The challenge here is that existing 3D training architecture usually lacks the ability to generalize well against rotated data. A natural solution could have been introducing rotation augmentations. However, observations show that existing architectures failed to benefit much from augmentations owing to the optimization difficulty and limited capacity of the network. Even worse, training with augmented data also introduces adversary effect that hurts the inference performance when test data is not rotated. Many works \cite{poulenard2019effective,rao2019spherical,kim2020rotation,yu2020deep,li2021closer,li2021rotation,zhang2019rotation,chen2019clusternet} attempted to address this issue through constructing rotation-invariant frameworks and features. Nevertheless, it is shown that rotation invariant features can still suffer evident performance drop when test data is inherently not rotated.   

\begin{figure*}[tb]
   \centering {\includegraphics[width=0.70\textwidth]{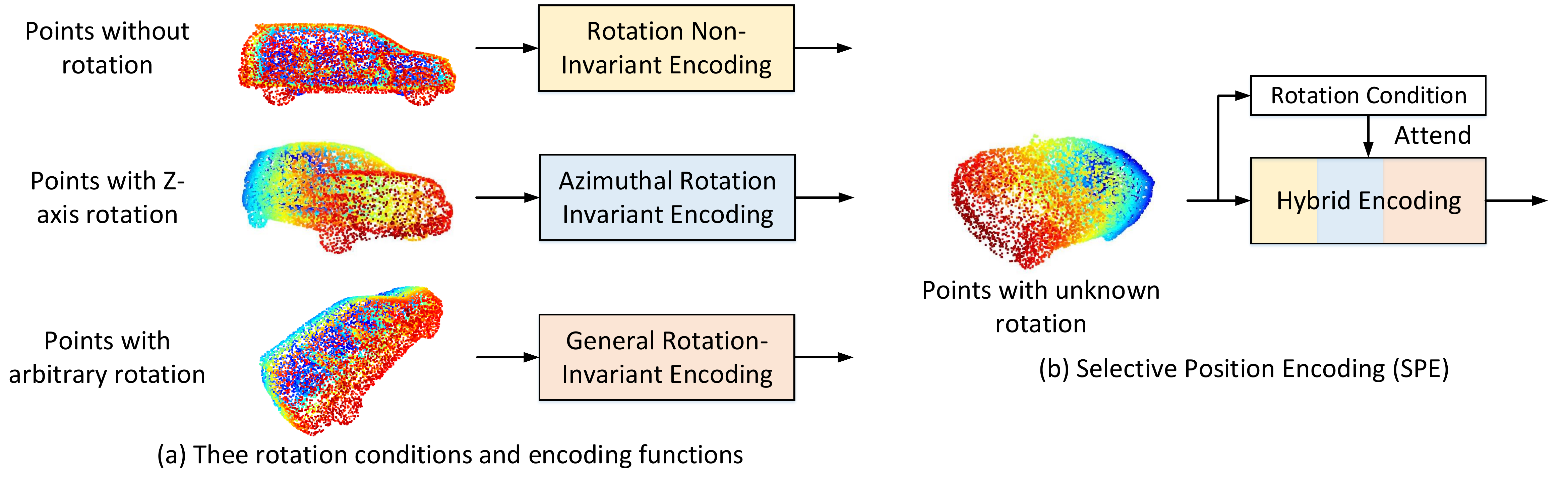}}
   \caption{Illustration of three different rotation conditions that SPE-Net models.}
   \label{fig:intro}
\end{figure*}

It therefore forms our main motivation to seek a training strategy that can capture the rotation information and adaptively adjust the parameter optimization based on such rotation condition. We illustrate the different types of rotation conditions considered in this paper. Fig. \ref{fig:intro}. (a) top is an object without rotation; Fig. \ref{fig:intro}. (a) middle illustrates an object that only exhibits Z-axis rotation;  Fig. \ref{fig:intro} (a) bottom demonstrates the general scenario where an object is rotated arbitrarily along all X-Y-Z axes. Why knowing the rotation information can be critical? A loose analogy perhaps can be the modeling through some conditional distribution $p(\mathbf\theta|r)$ versus the marginal distribution $p(\mathbf\theta)$, where the uncertainty of variable $\mathbf\theta$ always satisfies $\mathbb{E}[{\text{Var}}(\mathbf\theta|r)] \le {\text{Var}} (\mathbf\theta)$ for any random variables $\mathbf\theta$ and $r$. That being said, expected value of uncertainty on $\theta$ given observed evidence on $r$ would be guaranteed to reduce in comparison to total variance. If we view $\mathbf\theta$ as deep parameters, $r$ as rotation condition, the optimization on $\mathbf\theta$ can hopefully be restricted and eased given prior knowledge $r$.  

The design of SPE-Net in this paper devotes to the motivation described above. SPE-Net aims to learn a compatibility function that can attend to the actual rotation condition of the input. The training then enjoys a reduced degree of freedom of the optimization based on such compatibility function, where the rotation condition serves as useful prior knowledge, such as the $r$ variable. SPE-Net can henceforth flexibly benefit from stronger rotation training variations, without sacrificing much extra network capacity on encoding such global variations. In the meanwhile, SPE-Net effectively spares the learning to focus on finer local geometry structures, leading to better discriminative ability.

Essentially, the core SPE training unit consists of a critical attention mechanism and the special construction of ``position encoding features''.  In order to abstract useful rotation conditions without exposure to rotation annotations, SPE-Net jointly incorporates those three types of features that are sensitive to different positions: rotation non-invariant features (CD), azimuthal rotation invariant features (Z-RI), and the general rotation-invariant features (A-RI). This logic is briefly illustrated in Fig. \ref{fig:intro} (b). As the training proceeds in fitting the ground truths, the attention block gradually learns to capture the compatibility response between a point and each of those three types of features, i.e., CD, Z-RI or A-RI. Such learned attention then translates into the desired rotation information that SPE-Net can later leverage on. SPE-MLP, i.e., the critical building component of SPE-Net, is then enabled to adaptively discriminate and to attend to the relevant rotation condition, using this prior knowledge to focus on finer local geometry structures learning. This would potentially relieve the overall optimization difficulties, as the training is effectively constrained with restricted parameterization. Fig. \ref{fig:framework} illustrates the overview of the SPE-Net architecture. 

In brief summary, our contribution in this paper includes:

(1) We establish a new 3D point cloud training architecture, named SPE-Net. The proposed SPE-Net is a novel paradigm that can efficiently encode the rotation information of the input. This new architecture thus exhibits strong robustness against rotated input both during training and testing. 

(2) Further inspection into the SPE-Net reveals intriguing insights: We found it beneficial to always incorporate the rotation-invariant features properly during training, i.e., even if the test data does not exhibit any inherent global rotation. We envision that rotation can take place in local regions. SPE-Net thus exclusively benefits from a finer abstraction of both local and global rotation information, leading to superior robustness against variations.
    
(3) We demonstrate the strong empirical advantage of SPE-Net over SOTA methods. On challenging benchmarks such as ShapeNet, PartNet and S3DIS, SPE-Net achieve more than $1\%$ improvement, justifying the benefit through enhanced robustness against rotation.  

\section{Related Work}
\textbf{Deep learning for point cloud analysis.} The research in this direction has proceeded along two different dimensions: projecting the point cloud or using the original point cloud. For the first dimension, the original point clouds are projected to intermedia voxels \cite{zhou2018voxelnet,maturana2015voxnet,shi2020pv} or images \cite{you2018pvnet,li2020end}, translating the challenging 3D analysis into the well-explored 2D problem. These works avoid the direct process of irregular and unordered point cloud data, and show great efficiency benefited from the highly optimized implementation of convolutions. Nevertheless, Yang \emph{et al.} \cite{yang2019std} highlight the drawback of the projection step that loses the detailed information in point cloud, which limits the performances of the subsequent deep models. To overcome this limitation, the works along the second dimension utilize deep neural network to process the original point cloud. The pioneering work PointNet \cite{qi2017pointnet} proposes to model unodered point data through shared MLPs, and then is extended to PointNet++ \cite{qi2017pointnet++} by learning hierarchical representations through point cloud down-sampling and up-sampling. There are variants of approaches arisen from this methodology, which mainly focus on improving the capture of local geometry. For example, convolutions \cite{xu2021paconv,zhang2019shellnet,hermosilla2018monte,wu2019pointconv,thomas2019kpconv,li2018so,li2018pointcnn,su2018splatnet,xu2018spidercnn}, graph models \cite{lin2021learning,wang2019dynamic,shen2018mining,liu2019relation,zhao2019pointweb} and transformers \cite{zhao2021point,guo2021pct} are utilized to model local relation between neighboring points.

\noindent
\textbf{Rotation-robust networks.}
To improve the robustness of the network for rotation, a series innovations have been proposed to build rotation-equivariant networks or rotation-invariant networks. The rotation-invariant networks \cite{esteves2018learning,liu2018deep,shen20203d,rao2019spherical,weiler20183d} are required to produces the output features that are rotated correspondingly with the input. For example, Spherical CNN \cite{esteves2018learning} proposes to project the point cloud into spherical space and introduces a spherical convolution equivariant to input rotation. 3D steerable CNN \cite{weiler20183d} transfers the features into unit quaternions, which naturally satisfied the rotation-equivariant property. Another direction to enhance rotation-robustness is to learn rotation-invariant representations in networks. To achieve this, the rotation-invariant networks are derived by using kernelized convolutions \cite{poulenard2019effective,rao2019spherical}, PCA normalization \cite{kim2020rotation,yu2020deep,li2021closer} and rotation-invariant position encoding \cite{zhang2019rotation,li2021rotation,chen2019clusternet}.

Our work also falls into the category of rotation-robust architecture for point cloud analysis. Unlike the aforementioned methods that are required to guarantee the rotation-equivariant or rotation-invariant property, SPE-Net predicts the rotation condition based on the training data and adaptively chooses the optimal position encoding function. Therefore, this design, on one hand, improves the rotation robustness when the training data is manually rotated, and on the other, keeps the high learning capacity when rotation invariance is not required.

\section{SPE-Net}

We elaborate our core proposal: the Selective Position Encoding Networks (SPE-Net) in this section. The basic motivation is, if we have the knowledge of another random variable, i.e., the rotation condition $r$ of the input, we can use this knowledge to potentially reduce the expected uncertainty of the parameterization. This is because $\text{Var}(\theta)= {\mathbb{E}} ( \text{Var}(\theta|r))+ \text{Var}({\mathbb{E}}(\theta|r))$ for arbitrary $\theta$ and $r$. Note ${\mathbb{E}}[ \theta|r]$ is a function of $r$ by definition. If the three constructed features CD, Z-RI, and A-RI measurements can respond drastically different to the actual rotation condition $r$ of the input, showing different prediction behaviors, then the value $\text{Var}({\mathbb{E}}(\theta|r))$ would likely to be non-zero in our 3D context. This then leads to reduced uncertainty on the parameter estimation $\theta$, i.e., $\text{Var}(\theta) \ge {\mathbb{E}} ( \text{Var}(\theta|r))$. That forms a principled incentive of our SPE-Net to ease the optimization and to improve the prediction given prior knowledge on rotation condition $r$. 

\begin{figure*}[tb]
   \centering {\includegraphics[width=0.70\textwidth]{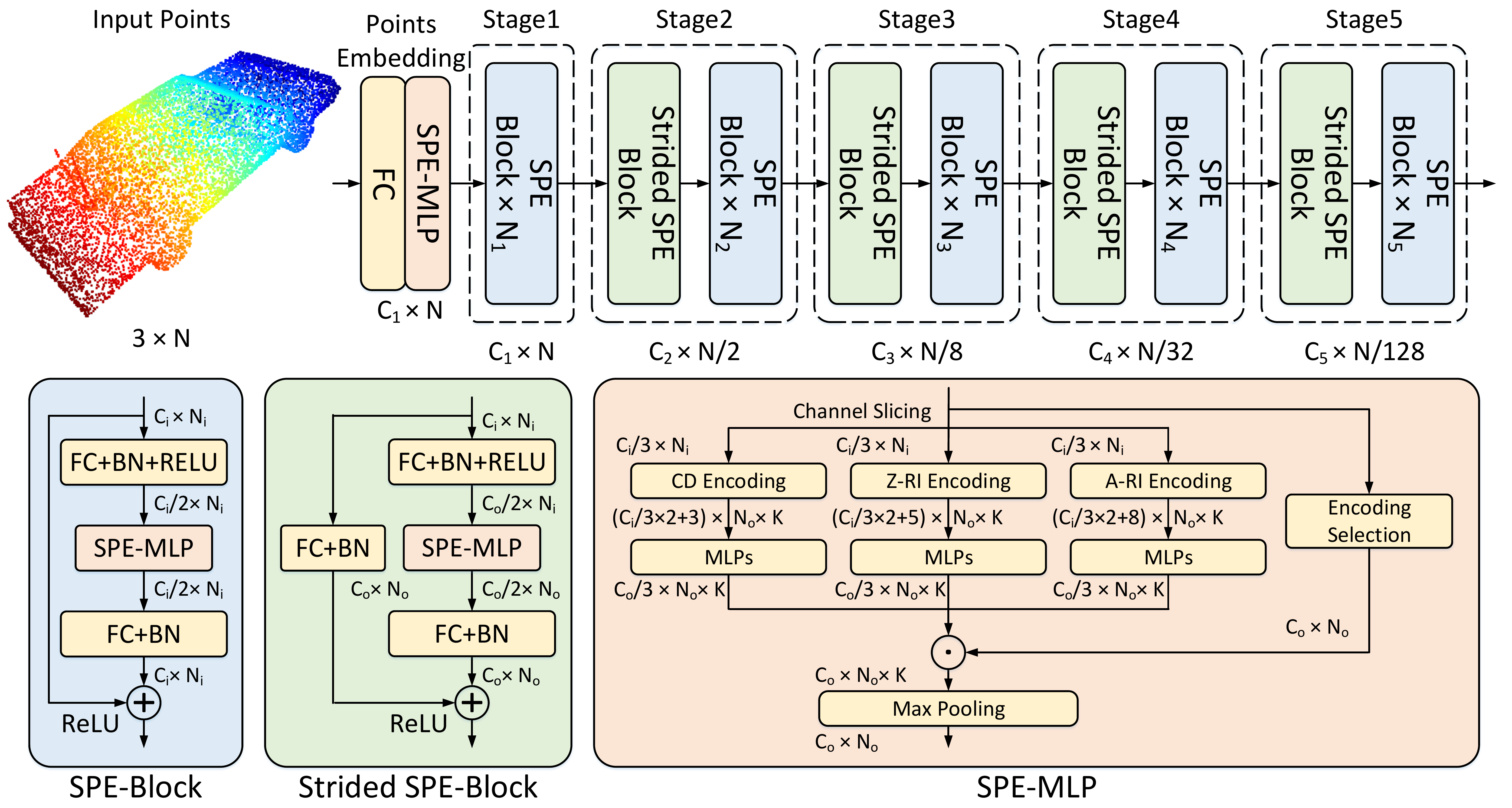}}
   \caption{An overview of our proposed Selective Position Encoding Networks (SPE-Net). It mainly consists of five stages of each stacks several SPE-Blocks. The first SPE-Block in each stage, called strided SPE-Block, increases the channel dimension while reduces the number of points. The size of output feature map is given as $num\_of\_channels \times num\_of\_points$ for each stage.
}
   \label{fig:framework}
\end{figure*}

\subsection{Overall Architecture Flow} \label{sec:overall}
The upper part of Fig. \ref{fig:framework} illustrates an overview of the SPE-Net. The basic architecture construction follows the philosophy of CNNs \cite{simonyan2014very,he2015deep}, where the channel dimension increases while the point number reduces and the layer goes deeper.

\textbf{Points embedding.} The 3D point cloud input is of size $3\times N$, where $N$ denotes the number of input points. SPE-Net firstly takes the input and embeds each query point with the information from its $K$-nearest neighboring points. In detail, the 3-dimensional coordinate of each point is firstly mapped into a feature $\mathbf{f}_{i}$ of higher dimension $C_1>3$ by using a shared linear embedding layer. An SPE-MLP operation then applies on the input $\mathbf{f}_{i}$ feature to further encapsulate the context feature from its neighboring points. Such point embedding module finally outputs features with a shape of $C_1 \times N$.

\textbf{Multi-stage architecture.} After the point embedding, several subsequent SPE-Blocks operate on the embedding out of the SPE-MLP layer. The overall structure can be grouped into five sequential stages, as illustrated in Fig. \ref{fig:framework}. In each stage, the building block named Strided SPE-Block increases the number of channels and down-samples the points. The feature resolutions are preserved in the following SPE-Blocks within the same stage. The number of SPE-Blocks $\{N_{1\sim 5}\}$ and channel dimensions $\{C_{1\sim 5}\}$ are considered as predefined hyperparameters that can be tailored for different point cloud analysis of different scales.

\textbf{Residual block.} Each residual block is composed of a shared fully-connected layer, an SPE-MLP operation and another shared fully-connected layer. The two fully-connected layers are utilized to respectively reduce and to recover the channel dimension, which behave similarly to the bottleneck structure in ResNet. Batch normalization is applied after each fully-connected layer.

\subsection{SPE-MLP} \label{sec:techlast}
Here we introduce the Selective Position Encoding MLP (SPE-MLP), our core learning unit. SPE-MLP builds upon traditional ``point-wise MLP'' structure while it provides further unique advantage in optimization from more restricted region of parameterisation. SPE-MLP can effectively predict the rotation condition $r$ of the input, and using this knowledge to both ease the training and improve the test accuracy.

\begin{figure*}[tb]
   \centering {\includegraphics[width=0.70\textwidth]{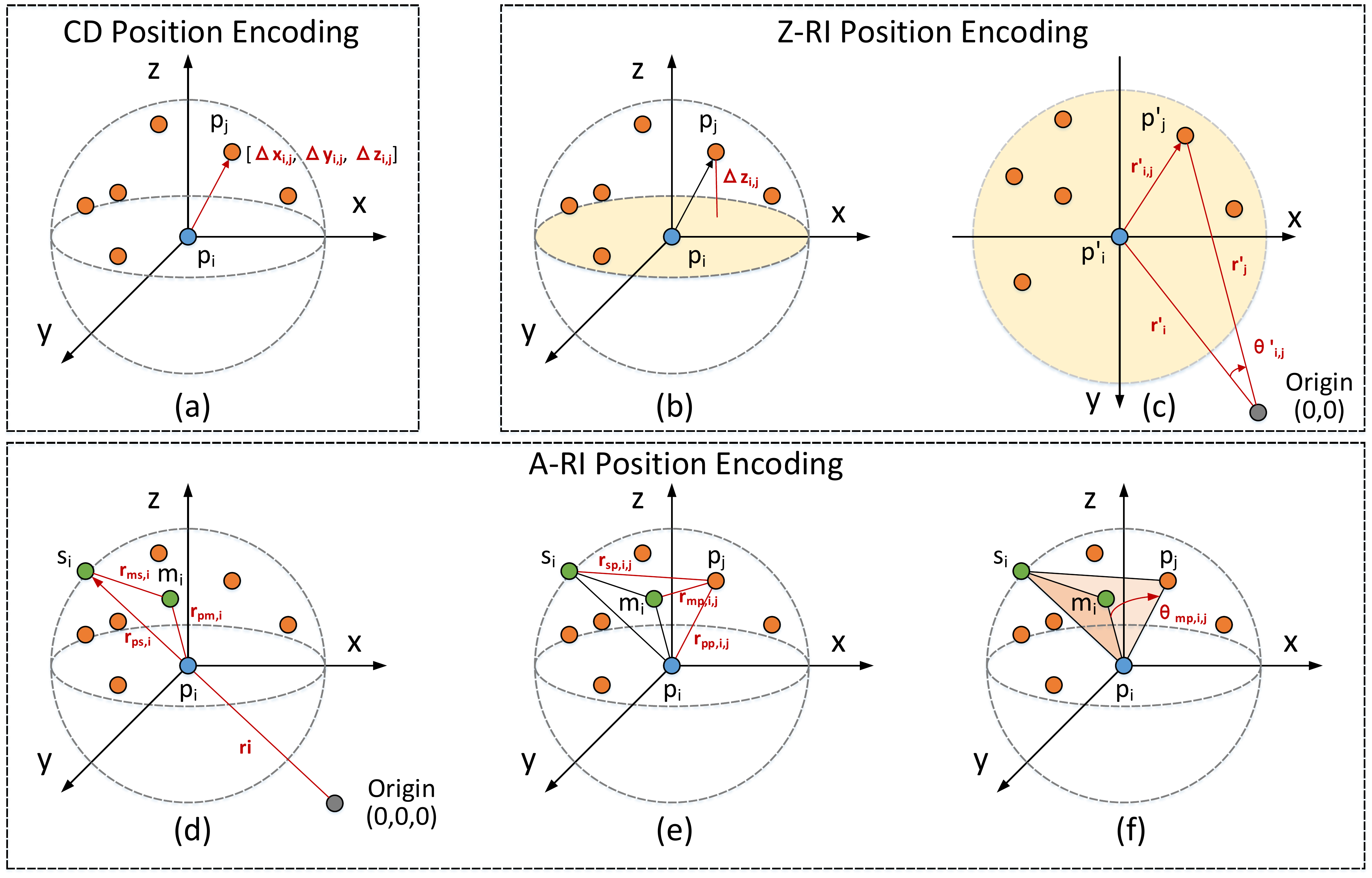}}
   \caption{An illustration of three position encoding functions including CD position encoding (a), Z-RI position encoding (b,c) and A-RI position encoding (d, e, f). The \textcolor[RGB]{0, 106, 255}{\textbf{blue}}, \textcolor[RGB]{255, 132, 0}{\textbf{orange}} and \textcolor[RGB]{8, 201, 72}{\textbf{green}} points are the query, neighboring and support points, respectively.}
   \label{fig:position_encoding}
\end{figure*}

\textbf{Position feature encoding.} SPE-MLP operates by computing a compatibility function between each query and a couple of constructed position encoding functions applied on the query. The hypothesis here is that each constructed position feature encoding function can respond discriminatively towards different rotation conditions of training data, thus effectively revealing a point's dependency with certain types of rotation condition. While the training proceeds, the compatibility function gradually learns to attend to the relevant position encoding functions according to how well such attention $r$ can reduce the total training risks. The attention information would then efficiently convert to the potential rotation condition $r$ of the input. We define three such position encoding functions that is practically suitable for our hypothesis, shown as in Fig.\ref{fig:position_encoding}. These constructed position encoding functions are described as following.

(a) Coordinate difference (CD) encoding. In the PointNet-style networks, the function called ``coordinate difference'' is usually exploited as the position encoding function. Formally, given the coordinate of query point $\mathbf{p}_i$ and its neighborhood $\mathbf{p}_{j}$, the CD position encoding is calculated by
\begin{equation}\label{eq:dc}
\begin{aligned}
P_{\text{CD}}(\mathbf{p}_i, \mathbf{p}_j)=\mathbf{p}_j-\mathbf{p}_i=[\Delta x_{i,j}, \Delta y_{i,j}, \Delta z_{i,j}]
\end{aligned}.
\end{equation}
CD position encoding is straightforward and shows promising performances in \cite{qi2017pointnet,qi2017pointnet++,wang2019dynamic,liu2020closer}. However, such encoding strategy generalizes poorly against rotation variations, since the feature construction is non-invariant against rotation variations. The phenomenon is also reported in recent works \cite{li2021rotation,xu2021sgmnet,chen2019clusternet,li2021closer,zhang2019rotation}. Nevertheless, this phenomenon forms basic evidence that supports our hypothesis: the objects having actual rotations indeed would inherently show much weaker response on CD features during training.

(b) Z-axis rotation-invariant (Z-RI) encoding. The second feature encoding function must be complementary to the above CD feature function so that the desired attention mechanism is discriminative. We firstly adopt the most simple heuristic that 3D objects usually merely exhibit rotation around the Z-axis (azimuthal rotation). In Eq.(\ref{eq:dc}), only the coordinate difference $\Delta z_{i,j}$ along Z-axis is Z-axis rotation-invariant. In order to better encode the relative position information along the other two axes, we further project the query point and its neighboring points to X-Y 2D plane, and utilize the distances between projected query $p'_i$, neighboring point $p'_{i,j}$ and origin point, along with the angle $\theta'_{i,j}$ as the encoded representation. Specifically, the Z-RI position encoding is formulated as
\begin{equation}\label{eq:zri}
\begin{aligned}
P_{\text{Z-RI}}(\mathbf{p}_i, \mathbf{p}_j)=[\Delta z_{i,j}, r'_{i,j}, r'_{i}, r'_{j}, \theta'_{i,j}]
\end{aligned}.
\end{equation}
The construction of the Z-RI position feature encoding is intended to show best compatibility response with the training data having azimuthal rotation, while showing weaker response towards alternative rotation conditions.   

(c) Arbitrary rotation-invariant (A-RI) encoding. The most general yet complex scenario is when arbitrary rotation takes place in 3D space. Compared to Z-axis rotation (azimuthal rotation), arbitrary rotation naturally incurs more complexity in parameterization using $\theta$. It is also challenging to construct arbitrary rotation-invariant representation with only query point and its neighboring points. Therefore, by following \cite{li2021rotation}, we additionally include two support points, i.e., the center point of neighborhoods $m_i$ and the intersection $s_i$ between the query ball and line extended from origin to query point. The center point is defined as the neighboring point having the minimum averaged distance from the other neighboring points. Eventually, A-RI position encoding consists of the distances among query, support and neighboring points, and the angle from $s_i-p_i-m_i$ plane to $s_i-p_i-p_{i,j}$ plane:
\begin{equation}\label{eq:ari}
\begin{aligned}
P_{\text{A-RI}}(\mathbf{p}_i, \mathbf{p}_j)=[r_i, r_{ms, i}, r_{ps, i}, r_{pm, i}, r_{sp, i, j}, r_{mp, i, j}, r_{pp, i, j}, \theta_{mp, i, j}]
\end{aligned}.
\end{equation}
The construction of the arbitrary rotation-invariant (A-RI) encoding is expected to exhibit the strongest dependency on arbitrarily rotated training data.

\textbf{Encoding Selection.} SPE-Net can dynamically attend to the relevant position encoding functions for each point $\textbf{p}_{i}$. This is realized through the construction of compatibility function as follows. For every stage of the SPE-Net, we firstly equally slice its input feature $\mathbf{f}_{i}$ into three partitions along channel dimension as $\mathbf{f}_i^{(1)},\mathbf{f}_i^{(2)},\mathbf{f}_i^{(3)}\in \mathbb{R}^{C/3}$. Three MLP mapping functions $F_1$, $F_2$ and $F_3$ are respectively applied to the features obtained via Eq. (\ref{eq:dc}),  (\ref{eq:zri}), (\ref{eq:ari}). Since the features Eq. (\ref{eq:dc}), (\ref{eq:zri}), (\ref{eq:ari}) intrinsically appear to exhibit different responses to an object's rotation condition, related to how well each option can reduce the actual training risk, the three MLP mapping functions $F_1$, $F_2$ and $F_3$ are expected to learn to abstract such preference during the training. The output of encoded local feature by using different position encoding is defined as:
\begin{equation}\label{eq:es1}
\begin{aligned}
&\textbf{g}_{i,j}^{(1)} = F_1([\mathbf{f}_{i}^{(1)}, \Delta\mathbf{f}_{i,j}^{(1)}, P_{\text{CD}}(\mathbf{p}_i, \mathbf{p}_j)]), \\
&\textbf{g}_{i,j}^{(2)} = F_2([\mathbf{f}_{i}^{(2)}, \Delta\mathbf{f}_{i,j}^{(2)}, P_{\text{Z-RI}}(\mathbf{p}_i, \mathbf{p}_j)]), \\
&\textbf{g}_{i,j}^{(3)} = F_3([\mathbf{f}_{i}^{(3)}, \Delta\mathbf{f}_{i,j}^{(3)}, P_{\text{A-RI}}(\mathbf{p}_i, \mathbf{p}_j)]). \\
\end{aligned}
\end{equation}
We calculate the weighted concatenation of the three encoded features followed by a max pooling layer:
\begin{align}
&[\alpha_{i}^{(1)}, \alpha_{i}^{(2)}, \alpha_{i}^{(3)}] = \text{Sigmoid}(\text{FC}(\mathbf{f}_{i})), \label{eq:es3} \\
& \textbf{g}_{i} = \mathop{\text{Max}}_{j}([\textbf{g}_{i,j}^{(1)} \odot \alpha_{i}^{(1)}, \textbf{g}_{i,j}^{(2)} \odot \alpha_{i}^{(2)}, \textbf{g}_{i,j}^{(3)} \odot \alpha_{i}^{(3)}]). \label{eq:es3b}
\end{align}
Here, the $\mathop{\text{Max}}$ operation serves as a maxing pooling. It outputs a vector that collects the entry-wise maximum value across all input vectors indexed by $j$. Symbol $\odot$ denotes element-wise product between two vectors that preserves the dimension of vectors. Weights $\alpha^{(1)}_{i}, \alpha^{(2)}_{i}, \alpha^{(3)}_{i}\in \mathbb{R}^{C/3}$ are learnable channel-wise attention vectors produced by an FC layer. These weights can learn to align with the respective $F_1$, $F_2$ and $F_3$ functions. Therefore, the training can be steered to focus on the relevant position encoding functions of each input, under the specific recognition task. All the weights are normalized to $0\sim 1$ by the sigmoid function. The whole process is analogous to softly selecting rotation conditions $r$ prior to the actual inference, so we call this procedure ``Encoding Selection''. 

Eq. (\ref{eq:es1}) is inspired from the Point-wise MLP structure, which was originally proposed in PointNet/PointNet++ \cite{qi2017pointnet,qi2017pointnet++}. A conventional Point-wise MLP applies several point-wise fully-connected (FC) layers on a concatenation of ``relative position'' and point feature to encode the context from neighboring points: 
\begin{equation}\label{eq:pw_mlp}
\begin{aligned}
\mathop{\text{Max}}_{j}(F([\mathbf{f}_{i}, \Delta\mathbf{f}_{i,j}, P(\mathbf{p}_i, \mathbf{p}_j)]))
\end{aligned},
\end{equation}
The max pooling operation across the $K$-nearest neighbors after the FC layer aggregates the feature of the query point, where $P(\cdot, \cdot)$ denotes the relatively position encoding between query point and neighboring point, $F(\cdot)$ is a point-wise mapping function (i.e., MLPs). The point-wise MLPs after concatenation operation can approximate any continuous function about the relative coordinates and context feature \cite{qi2017pointnet++}. They also utilize the ball radius method \cite{qi2017pointnet++} to achieve $K$ neighboring points $\{\mathbf{p}_{j}, \mathbf{f}_{j}\}$, which is critical to achieve balanced sample density. Eq. (\ref{eq:es1}) borrow all of these merits from Point-wise MLP, while Eq. (\ref{eq:es1}) exclusively incorporates the rotation information to achieve our goal. 

Apart from the learnable weights, good initialization can help further improve the rotation robustness of SPE-Net. Particularly, we can mask out the CD and Z-RI position encoding in Eq(\ref{eq:es1}) during the first $T$ epochs. In this way, the networks are forced to only use rotation-invariant features (A-RI) to represent the 3D object, at the earlier training stage. After $T$ epochs, the network starts to incorporate the information from the other two position encodings (CD and Z-RI). The mask-out epochs $T$ here is a trade-off hyperparameter. A much higher $T$ will increase the rotation robustness of networks whereas it might hurt the performance due to the lack of training flexibility. We conduct ablation study against $T$ in the experiments.

\section{Experiments}
\subsection{Datasets}
We empirically evaluate our SPE-Net on four challenging point cloud analysis datasets: ModelNet40 \cite{wu20153d} for 3D classification, ShapeNet \cite{chang2015shapenet}, PartNet \cite{mo2019partnet} for part segmentation and S3DIS \cite{armeni2017joint} for scene segmentation. The first three datasets are generated from 3D CAD models, and the last one is captured in real scenes.

\textbf{ModelNet40} is one of the standard benchmarks for 3D classification. The dataset contains 12.3K meshed CAD models from 40 classes. The CAD models are officially divided into 9.8K and 2.5K for training and testing, respectively.

\textbf{ShapeNet} is a standard part segmentation benchmark, covering 16 categories and 50 labeled parts. It consists of 12.1K training models, 1.9K validation models and 2.9K test models. We train our model on the union of training set and validation set, and report the performances on the test set.

\textbf{PartNet} is a more challenging benchmark for large-scale fine-grained part segmentation. It consists of pre-sampled point clouds of 26.7K 3D objects from 24 categories, annotated with 18 parts on average. We follow the official data splitting scheme, which partitions the objects into 70\% training set, 10\% validation set and 20\% test set. We train our model on the training set and report the performances on validation set and test set, respectively.

\textbf{S3DIS} is an indoor scene segmentation dataset captured in 6 large-scale real indoor areas from 3 different buildings. There are 273M points in total labeled with 13 categories. Following \cite{tchapmi2017segcloud}, we use Area-5 as the test scene and the others as training scenes. Considering that each scene contains a large amount of points exceeding the processing capacity of GPU device, for each forward, we segment sub-clouds in spheres with radius of 2m. Such sub-clouds are randomly selected in scenes during training, and evenly sampled for testing.

\subsection{Implementation Details}
\noindent
\textbf{Backbone network architectures.}
As described in Sec. \ref{sec:overall}, the complexity of SPE-Net is determined by the settings of free parameters. We fix the repeat number as $N_t=1|_{1\leqslant t\leqslant 5}$, which means each stage contains one SPE-Blocks. The number of MLP layers in SPE-MLP block is set as $1$. The number of output channels $C_t$ is tuned to build a family of SPE-Net with various model complexities. For small-scale ModelNet40 and ShapeNet, we set $\{C_t\}=\{72, 144, 288, 576, 1152\}$, which are then expanded by 2$\times$ for large-scale PartNet and S3DIS.

\noindent
\textbf{Head network architectures.}
We attach a head network on top of SPE-Net to train for different tasks. For the classification head, the output features of SPE-Net are aggregated together by a max-pooling layer, followed by a three-layer MLP with output channels $576$-$288$-$c$ to perform $c$-class prediction. For the task of segmentation, we follow the architecture in \cite{qi2017pointnet++} that progressively upsamples the output features from the backbone network and reuses the features from the earlier layers by the skip connections.

\noindent
\textbf{Training and inference strategy.}
Our proposal is implemented on PyTorch \cite{NEURIPS2019_9015} framework, and SGD (for ModelNet40 and S3DIS) or AdamW (for ShapeNet and PartNet) algorithm is employed to optimize the model with random initialization. Considering that training deep models usually requires extensive data augmentation \cite{liu2020closer,li2021contextual,touvron2021training,long2022stand,qiu2022mlp}, we exploit dropout, drop path, label smoothing, anisotropic random scaling and gaussian noise to reduce the over-fitting effect. In the inference stage, we follow the common test-time augmentation of voting scheme \cite{qi2017pointnet++,liu2020closer,thomas2019kpconv} and augment each model 10 times using the same augmentation strategy in training stage.

\subsection{Ablation Study on SPE-MLP}
\begin{table}[!tb]
\centering
\scriptsize
\caption{Ablation study for each design in SPE-MLP on ModelNet40 dataset. The backbone network is SPE-Net-S.}
\begin{tabularx}{0.88\textwidth}{l|cccc|C|CCCC}
\toprule
{Method} & {CD} & {Z-RI} & {A-RI} & {Sel} & {param} & {N/N} & {Z/Z} & {Z/SO3} & {SO3/SO3} \\
\midrule
\multirow{3}{*}{Single type} &\checkmark & & & & 1.6M & \underline{93.6} & 92.0 & 22.3 & 90.8 \\
& &\checkmark & & & 1.6M & 92.9 & \underline{92.3} & 19.9 & 91.0 \\
& & &\checkmark & & 1.6M & 90.6 & 91.4 & \textbf{91.4} & \underline{91.3} \\ \midrule
\multirow{2}{*}{Multiple types} & \checkmark &\checkmark &\checkmark & & 1.3M & 93.5 & 92.1 & 26.0 & 90.7 \\
& \checkmark &\checkmark &\checkmark &\checkmark & 1.5M & \textbf{93.9} & \textbf{92.6} & \underline{88.4} & \textbf{91.5} \\
\bottomrule
\end{tabularx}
\label{tab:ablation}
\end{table}
We firstly study how each particular design in SPE-MLP influences the overall performances. The basic strategy is to utilize only a single type of position encoding from either CD, Z-RI or A-RI encoding functions and compare with our attention based learning. To test these variants, we do not slice the input channels, and instead, we concatenate the full input channels with the certain single type of position encoding. For those relevant architecture variants that involve multiple types of position encoding, the three position encoding functions are either simply combined or fused by the proposed encoding selection (abbreviated as Sel). We compare the models under consideration on four different rotation conditions:
(1) The original training and testing sets without any rotation: N/N.
(2) Both training and testing sets are rotated around the Z-axis: Z/Z.
(3) Training set is rotated around the Z-axis and testing set is randomly rotated: Z/SO3.
(4) Both training and testing sets are randomly rotated: SO3/SO3.
Table \ref{tab:ablation} evaluates the performances of different variants of SPE-MLP on ModelNet40 dataset. The ablation study aims to justify our basic hypotheses. For rapid comparisons, we exploit a lightweight version of SPE-Net, called SPE-Net-S, with output channels as $\{C_t\}=\{36, 72, 144, 288, 576\}$.

The first observation is intuitive and forms our basic motivation. When a single type of position encoding, i.e., one out of CD, Z-RI, or A-RI encoding is applied in the network, such selected encoding strategy would perform the best when the training data intrinsically exhibit such type of rotation. Take for instance, CD performs the best when both the training and test data shows no rotation. A-RI significantly outperforms other encoding schemes when the training data is rotated around the Z-axis while the test data is arbitrary rotaed. In the meanwhile, it is also apparent that when the position encoding scheme does not model certain type of invariance assumption, the training fails drastically in learning that type of rotation condition, such as the CD and Z-RI features against the arbitrary rotation observed in test data. However, it is also noted that even if we train the network with arbitrarily rotated invariance features (A-RI), such training can hardly outperform tailored feature encoding schemes, for example, the CD feature under N/N setup or Z-RI under the Z/Z setup. 

One step further is to rigidly fuse all types of position condition with equal weights without attention based adaptation. In this way, the network are at least encouraged to make the predication based on all possible rotation conditions. We notice that such rigid combination of position encoding features indeed shows evident performance boost when training data and test data shows consistent rotation conditions. However, such encoding scheme performs poorly whenever the the training and test data differs in their rotation type, owing to poor generalization of such encoding scheme.

In comparison, the proposed position encoding selection (the Sel.column in Table \ref{tab:ablation}) based on the constructed attention scheme shows best robustness against rotation conditions both during training and testing. This is because SPE-net can adaptively learn to use the learned rotation condition as important prior information to make the prediction. Among all the compared counterparts when the training data and test data of the same rotation conditions, Sel. (implemented via SPE-MLP) performs consistently the best. The Sel. scheme also only incurs very limited performance drop when training and testing data are of different rotation types, justifying the advantage of the SPE attention operation and the good generalization ability. The only exception is that Sel. performs relatively worse than the A-RI scheme under Z/SO3 setup. However, recall that the A-RI can hardly outperform tailored feature encoding schemes whenever training data rotation matches the particularly encoding invaraince assumption. A-RI feature also performs worse than Sel under all other training scenarios. 

\begin{figure}[!tb]
\centering
   \subcaptionbox{ }{
     \includegraphics[width=0.25\textwidth]{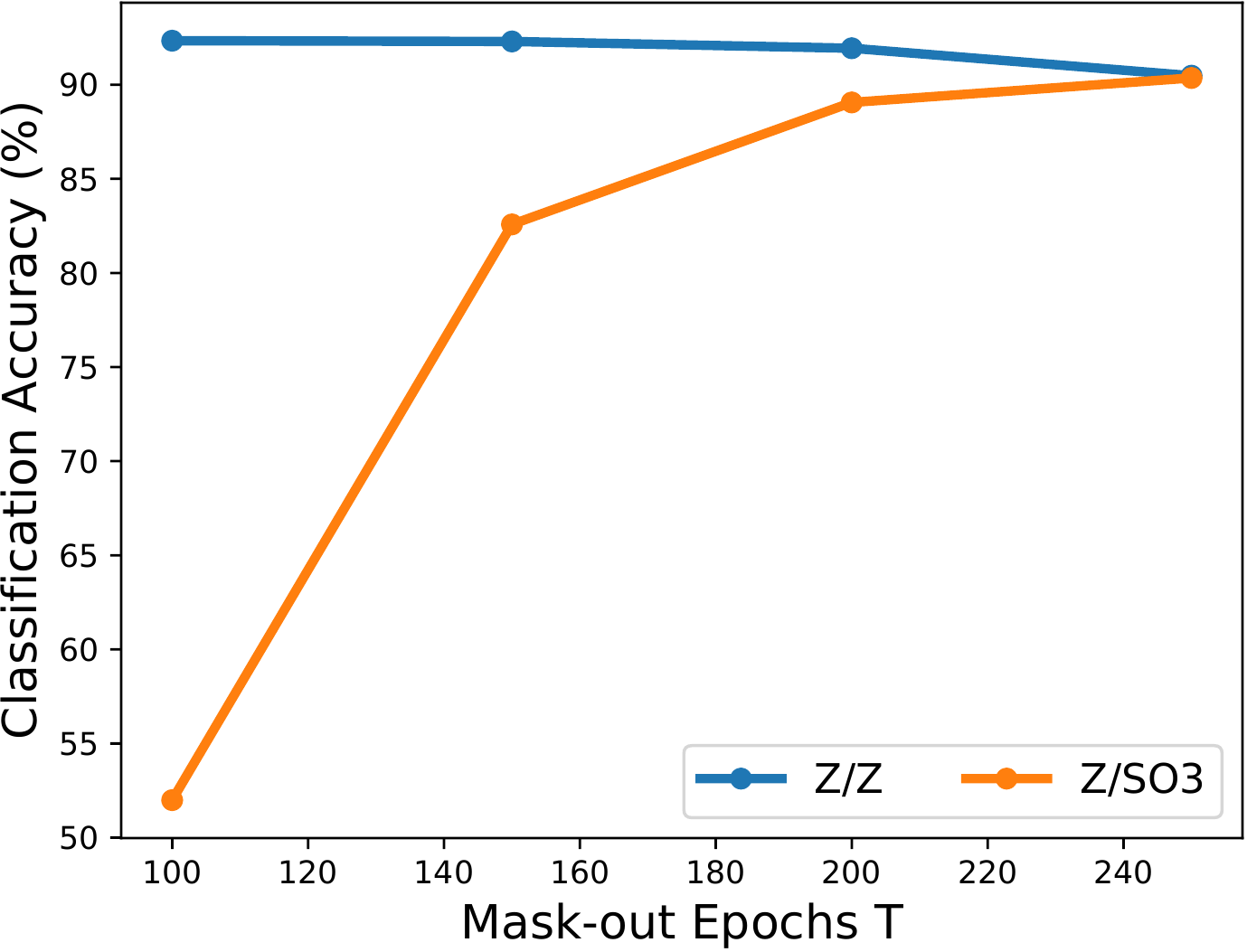}}
   \subcaptionbox{ }{
     \includegraphics[width=0.255\textwidth]{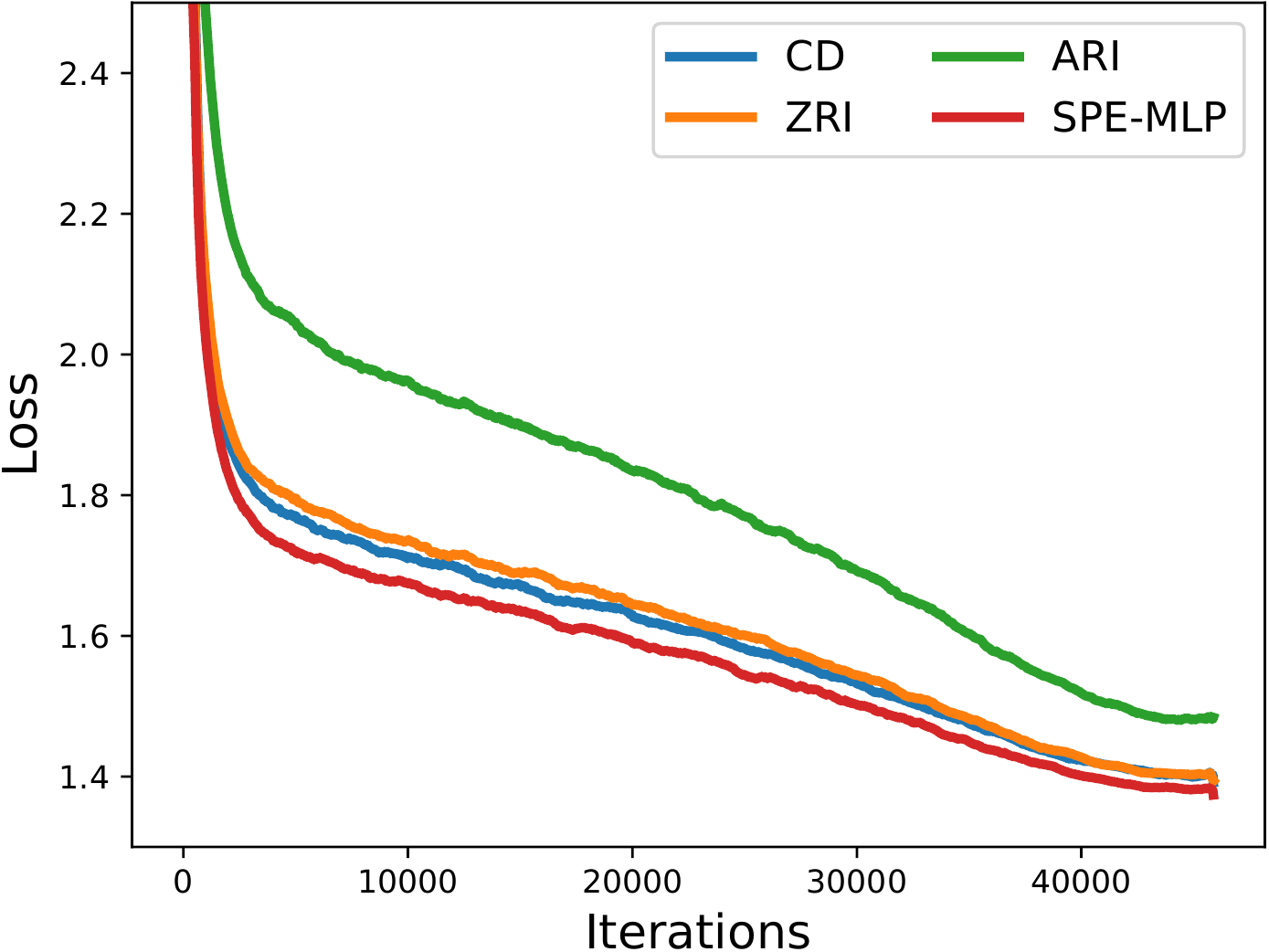}}
   \subcaptionbox{ }{
     \includegraphics[width=0.255\textwidth]{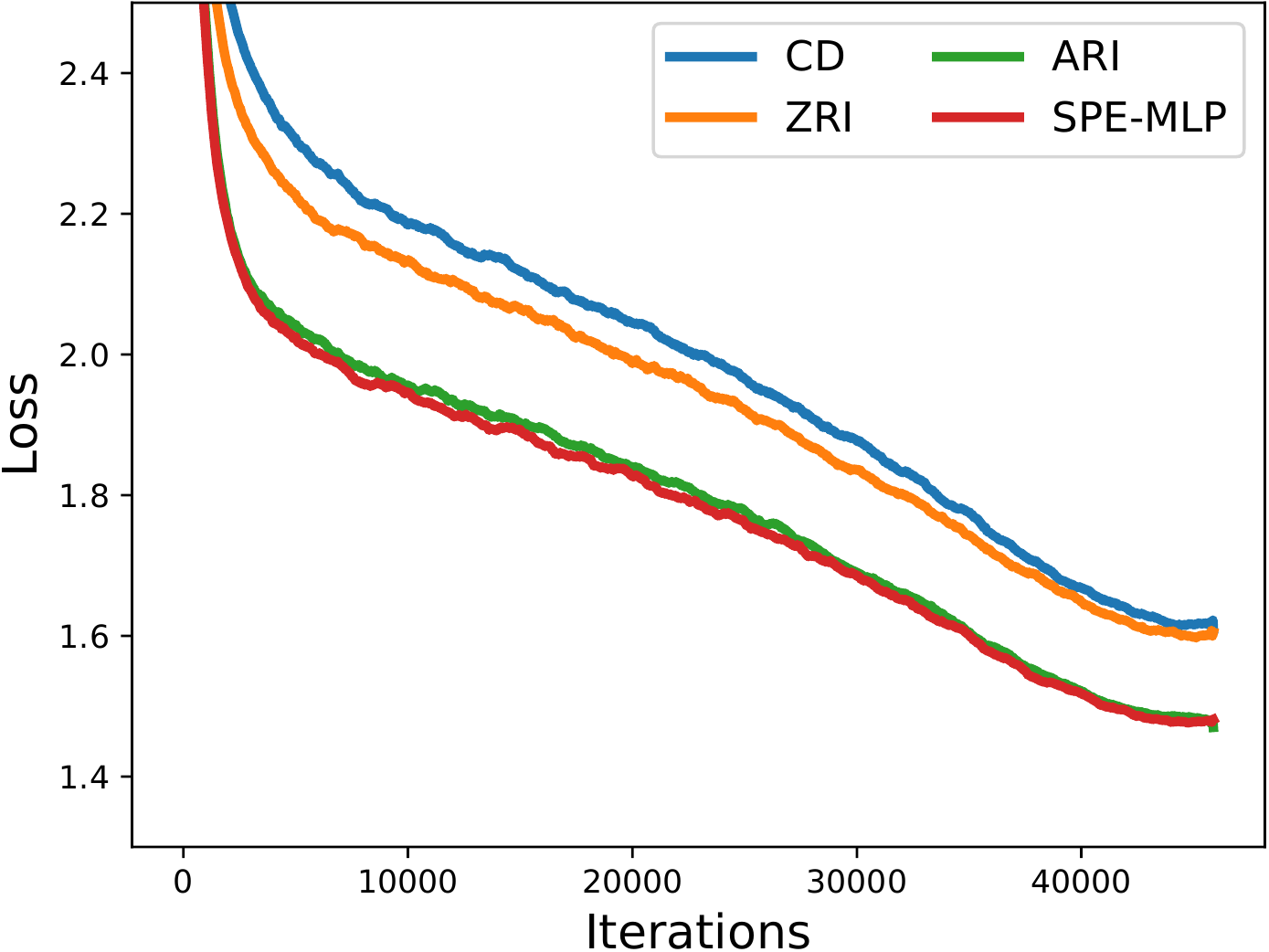}}
   \caption{(a) Effect of mask-out epochs. (b) Loss curves on ModelNet40 without rotation. (c) Loss curves on ModelNet40 with SO3 rotation.}
   \label{fig:curve}
\end{figure}

\begin{figure*}[tb]
   \centering {\includegraphics[width=0.85\textwidth]{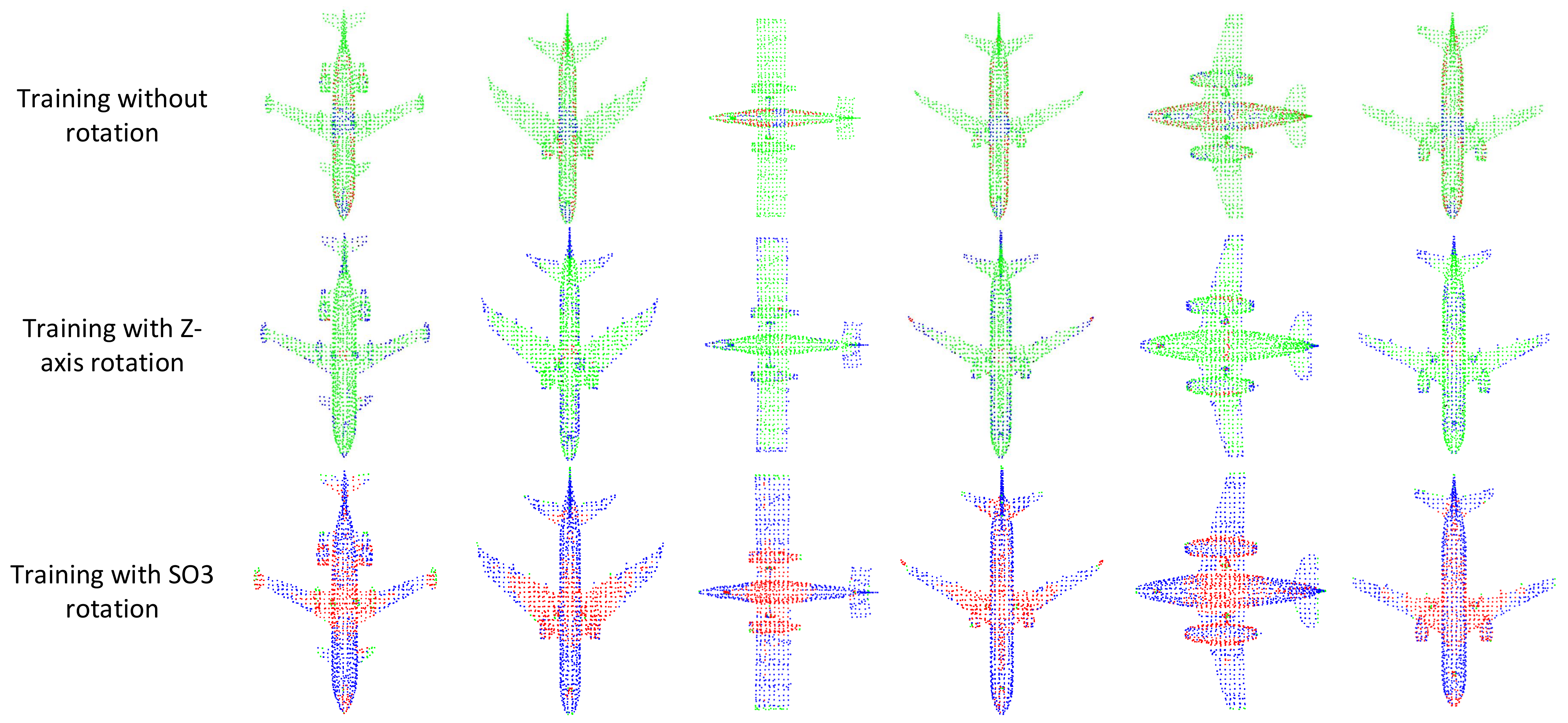}}
   \caption{The visualization of encoding selection on ModelNet40 dataset with different rotation condition. The \textcolor{red}{\textbf{red}}, \textcolor{green}{\textbf{green}} and \textcolor{blue}{\textbf{blue}} points represent the points dominated with CD, Z-RI and A-RI encoding, respectively.}
   \label{fig:visualization}
\end{figure*}

To clarify the effect of the mask-out epochs $T$ (Section \ref{sec:techlast}), we illustrate SPE-Net's performance curves under Z/Z and Z/SO3 setups on ModelNet40 against different mask-out epochs $T$ in Fig. \ref{fig:curve}(a). As can be seen, with higher $T$, the accuracy of Z/Z setting slightly drops while the transferring setting Z/SO3 is improved. We choose a good trade-off and set $T=200$ as the default choice. Fig. \ref{fig:curve}(b) and \ref{fig:curve}(c) respectively shows the loss curves on ModelNet40 without rotation or with SO3 rotation, and compares SPE-MLP against three other single-type position encoding. The losses of SPE-MLP are much lower than others across different iterations and different rotation conditions, and validate the robustness of SPE-MLP and the eased training optimization. Moreover, we also visualize the produced attention weights in Fig. \ref{fig:visualization}. The attention weights of ``encoding selection'' in the first strided SPE-Block are visualized. Overall, the models trained without rotation and that with Z-axis rotation prefer to choose Z-RI encoding condition, while the model trained with SO3 rotation utilizes more A-RI encoding condition. An intriguing observation is that, the same parts of a shape are very likely to attend to the same position encoding condition across different instances. It somewhat reveals that the proposed encoding selection achieves a unique fine-grained understanding in different parts of the 3D shape.

\subsection{Evaluation on Point Cloud with Rotation}
\begin{table}[!tb]
\centering
\scriptsize
\caption{Performances on ModelNet40 for 3D classification with different rotation conditions. The last column records the differences between Z/SO3 and SO3/SO3.}
\begin{tabularx}{0.92\textwidth}{l|c|c|CCCC}
\toprule
Type & Method & Inputs & {Z/Z} & {Z/SO3} & {SO3/SO3} & Acc.drop \\
\midrule
\multirow{4}{*}{Rotation-sensitive} & PointNet \cite{qi2017pointnet} & xyz & 88.5 & 16.4 & 70.5 & 54.1 \\
& PointNet++ \cite{qi2017pointnet++} & xyz & 89.3 & 28.6 & 85.0 & 56.4 \\
& DGCNN \cite{wang2019dynamic} & xyz & \underline{92.2} & 20.6 & 81.1 & 60.5 \\
& PointConv \cite{wu2019pointconv} & xyz & 91.6 & - & 85.6 & - \\
\midrule
\multirow{14}{*}{Rotation-robust} & Spherical CNN \cite{esteves2018learning} & voxel & 88.9 & 76.9 & 86.9 & 10.0 \\
& $\alpha^{3}$SCNN \cite{liu2018deep} & voxel & 89.6 & 87.9 & 88.7 & 0.8 \\
& RIConv \cite{zhang2019rotation} & feature & 86.5 & 86.4 & 86.4 & 0.0\\
& SRI-Net \cite{sun2019srinet} & feature & 87.0 & 87.0 & 87.0 & 0.0 \\
& SPH-Net \cite{poulenard2019effective} & xyz & 87.7 & 86.6 & 87.6 & 1.0 \\
& SFCNN \cite{rao2019spherical} & xyz & 91.4 & 84.8 & 90.1 & 5.3 \\
& PFE \cite{yu2020deep} & xyz+feature & 89.2 & 89.2 & 89.2 & 0.0 \\
& RI-GCN \cite{kim2020rotation} & xyz & 89.5 & 89.5 & 89.5 & 0.0 \\
& REQNN \cite{shen20203d} & xyz & 83.0 & 83.0 & 83.0 & 0.0 \\
& RotPredictor \cite{fang2020rotpredictor} & xyz & 92.1 & - & 90.8 & - \\
& Triangle-Net \cite{xiao2021triangle} & feature & - & - & 86.7 & - \\
& RIF \cite{li2021rotation} & feature & 89.4 & 89.4 & 89.3 & 0.1 \\
& DGCNN+PCA  \cite{li2021closer} & xyz & 91.6 & \textbf{91.6} & \underline{91.6} & 0.0 \\
\cmidrule{2-7}
& \textbf{SPE-Net (ours)} & xyz & \textbf{92.7} & \underline{89.7} & \textbf{91.8} & 2.1 \\
\bottomrule
\end{tabularx}
\label{tab:rotate}
\end{table}

Next, we compare with several state-of-the-art techniques on ModelNet40 and ShapeNet with different rotation conditions to demonstrate the advantage of the proposed SPE-Net. The performance comparisons on ModelNet40 for 3D shape classification are summarized in Table \ref{tab:rotate}. It is observed that rotation-sensitive methods such as PointNet and PointNet++ completely failed to generalize when test data shows different rotation than what it was trained on. The performance of these models also plummet under the SO3/SO3 setup owing to optimization difficulty raised by complex rotation augmentations. On the contrary, the SOTA rotation robust methods show certain robustness against rotations, particularly under the SO3/SO3 and Z/SO3 setting. But these methods perform generally worse than other paradigms, especially under the Z/Z setting. This is because the network parameterization had to trade-off the rotation robustness against the general prediction performance, under the limited network capacity. Our SPE-Net is also a kind of rotation robust 3D training scheme. Particularly, as SPE-Net further incorporates the rotation condition as prior, it can flexibly adapt to the most relevant position encoding schemes, showing both good robustness against rotations, as well as the eased optimization difficulties in terms of good generalization ability. To see this, SPE-Net reports the best $92.7\%$ accuracy under the standard Z/Z setup, while it also simultaneously preserves good generalization ability against rotations changes. In comparison to DGCNN+PCA, SPE-Net still maintains a $0.2\%$ superiority under the SO3/SO3 setup, proving the unique advantage of SPE-MLP encoding scheme. 

\begin{table}[!tb]
\centering
\scriptsize
\caption{Performances on ShapeNet for part segmentation with different rotation conditions. The last column records the differences between Z/SO3 and SO3/SO3.}
\begin{tabularx}{0.78\textwidth}{l|c|c|CCC}
\toprule
Type & Method & Inputs & {Z/SO3} & {SO3/SO3} & mIoU.drop \\
\midrule
\multirow{4}{*}{Rotation-sensitive} & PointNet \cite{qi2017pointnet} & xyz & 37.8 & 74.4 & 36.6 \\
& PointNet++ \cite{qi2017pointnet++} & xyz & 48.2 & 76.7 & 28.5 \\
& PointCNN \cite{li2018pointcnn} & xyz & 34.7 & 71.4 & 36.7 \\
& DGCNN \cite{wang2019dynamic} & xyz & 37.4 & 73.3 & 35.9 \\
\midrule
\multirow{7}{*}{Rotation-robust} & RIConv \cite{zhang2019rotation} & feature & 75.3 & 75.5 & 0.2 \\
& SRI-Net \cite{sun2019srinet} & feature & 80.0 & 80.0 & 0.0 \\
& RI-GCN \cite{kim2020rotation} & xyz & 77.2 & 77.3 & 0.1 \\
& Triangle-Net \cite{xiao2021triangle} & feature & - & 72.5 & - \\
& RIF \cite{li2021rotation} & feature & 82.2 & 82.5 & 0.3 \\
& DGCNN+PCA  \cite{li2021closer} & xyz & \underline{83.1} & \underline{83.1} & 0.0 \\
\cmidrule{2-6}
& \textbf{SPE-Net (ours)} & xyz & \textbf{87.1} & \textbf{87.8} & 0.7 \\
\bottomrule
\end{tabularx}
\label{tab:rotate2}
\end{table}

Table \ref{tab:rotate2} summarizes the performance comparisons on ShapeNet for 3D part segmentation task. For evaluation on ShapeNet, we calculate the mean intersection of union (mIoU, \%) for each shape and report the mean value over all instances. Generally speaking, we observe similar behaviors of various methods on ShapeNet as that in Table \ref{tab:rotate}. Here rotation-sensitive methods still perform poorly under Z/SO3 and SO3/SO3 setups. We also observe a more significant performance gap between SPE-Net and DGCNN+PCA than that in Table \ref{tab:rotate}. The reason might be that for segmentation tasks, different input points may need drastically different attention to each of the position encoding scheme, demonstrating a more local response towards the variation of the rotation conditions. SPE-Net can therefore benefit significantly from its more flexible and pointwise rotation condition adaptation to reach better segmentation results.  

\subsection{Evaluation on Point Cloud without Rotation}
\begin{table}[!tb]
\centering
\scriptsize
\caption{Performances on ModelNet40, PartNet and S3DIS without test-time rotation.}
\begin{tabularx}{0.88\textwidth}{l|CC|CCC|CC}
\toprule
\multirow{2}{*}{Method} & \multicolumn{2}{c|}{ModelNet40} & \multicolumn{3}{c|}{PartNet} & \multicolumn{2}{c}{S3DIS} \\
\cmidrule{2-8}
& param & acc & param & val & test & param & mIoU \\
\midrule
DensePoint \cite{liu2019densepoint} & 0.7M & \underline{93.2} & - & - & - & - & - \\
KPConv \cite{thomas2019kpconv} & 15.2M & 92.9 & - & - & - & 15.0M & 65.7 \\
PointCNN \cite{li2018pointcnn} & 0.6M & 92.5 & 4.4M & - & 46.4 & 4.4M & 65.4 \\
point-wise MLP \cite{liu2020closer} & 26.5M & 92.8 & 25.6M & 48.1 & 51.5 & 25.5M & 66.2 \\
pseudo grid \cite{liu2020closer} & 19.5M & 93.0 & 18.5M & 50.8 & 53.0 & 18.5M & 65.9 \\
adapt weights \cite{liu2020closer} & 19.4M & 93.0 & 18.5M & 50.1 & 53.5 & 18.4M & 66.5 \\
PosPool \cite{liu2020closer} & 19.4M & \underline{93.2} & 18.5M & \underline{50.6} & \underline{53.8} & 18.4M & \underline{66.7} \\ 
\midrule
\textbf{SPE-Net (ours)} & 7.2M & \textbf{94.0} & 24.4M & \textbf{52.6} & \textbf{54.8} & 24.3M & \textbf{67.8} \\
\bottomrule
\end{tabularx}
\label{tab:unrotate}
\end{table}

Finally, we compare with several state-of-the-art techniques on the original training/testing sets of ModelNet40, PartNet and S3DIS for 3D shape classification, 3D part segmentation, 3D scene segmentation, respectively, to further validate the effectiveness of SPE-Net as a general backbone. The performances are summarized in Table \ref{tab:unrotate}. SPE-Net apparently has achieved the best performance among all comparisons. Note that the conventional point-wise MLP only takes into account the CD encoding, thus leading to a much worse performance than SPE-Net, even if both of the networks have adopted the residual learning structures. It is also intriguing to see that SPE-Net performs more than $1.1\%$ better than its best competitor PosPool on the S3DIS dataset. Such phenomenon verifies that it is always beneficial to incorporate the rotation-invariant features properly during training, although the data does not exhibit any inherent global rotation. This is because rotation can take place in local regions. SPE-Net thus exclusively benefits from a finer abstraction of both local and global rotation information, showing consistent performance gain in comparison to others.     

\section{Conclusion}
In this paper, we investigate the role of rotation conditions in 3D application's performance, and propose a novel deep 3D architecture named SPE-Net. SPE-Net can learn to infer and to encode various rotation condition present in the training data through an efficient attention mechanism. Such rotation condition then can effectively serve as useful prior information in improving the eventual prediction performance. Extensive empirical results validate our various assumptions and well verify the advantage of SPE-Net by leveraging on these priors.     

\textbf{Acknowledgments.} This work was supported by the National Key R\&D Program of China under Grant No. 2020AAA0108600. 

%
%
\bibliographystyle{splncs04}

\begin{thebibliography}{10}
\providecommand{\url}[1]{\texttt{#1}}
\providecommand{\urlprefix}{URL }
\providecommand{\doi}[1]{https://doi.org/#1}

\bibitem{armeni2017joint}
Armeni, I., Sax, S., Zamir, A.R., Savarese, S.: Joint 2d-3d-semantic data for
  indoor scene understanding. arXiv preprint arXiv:1702.01105  (2017)

\bibitem{chang2015shapenet}
Chang, A.X., Funkhouser, T., Guibas, L., Hanrahan, P., Huang, Q., Li, Z.,
  Savarese, S., Savva, M., Song, S., Su, H., et~al.: Shapenet: An
  information-rich 3d model repository. arXiv preprint arXiv:1512.03012  (2015)

\bibitem{chen2019clusternet}
Chen, C., Li, G., Xu, R., Chen, T., Wang, M., Lin, L.: Clusternet: Deep
  hierarchical cluster network with rigorously rotation-invariant
  representation for point cloud analysis. In: CVPR (2019)

\bibitem{esteves2018learning}
Esteves, C., Allen-Blanchette, C., Makadia, A., Daniilidis, K.: Learning so (3)
  equivariant representations with spherical cnns. In: ECCV (2018)

\bibitem{fang2020rotpredictor}
Fang, J., Zhou, D., Song, X., Jin, S., Yang, R., Zhang, L.: Rotpredictor:
  Unsupervised canonical viewpoint learning for point cloud classification. In:
  3DV (2020)

\bibitem{guo2021pct}
Guo, M.H., Cai, J.X., Liu, Z.N., Mu, T.J., Martin, R.R., Hu, S.M.: Pct: Point
  cloud transformer. Computational Visual Media  \textbf{7}(2),  187--199
  (2021)

\bibitem{he2015deep}
He, K., Zhang, X., Ren, S., Sun, J.: Deep residual learning for image
  recognition. In: CVPR (2016)

\bibitem{hermosilla2018monte}
Hermosilla, P., Ritschel, T., V{\'a}zquez, P.P., Vinacua, {\`A}., Ropinski, T.:
  Monte carlo convolution for learning on non-uniformly sampled point clouds.
  ACM Trans. on Graphics  \textbf{37}(6),  1--12 (2018)

\bibitem{kim2020rotation}
Kim, S., Park, J., Han, B.: Rotation-invariant local-to-global representation
  learning for 3d point cloud. In: NeurIPS (2020)

\bibitem{li2021closer}
Li, F., Fujiwara, K., Okura, F., Matsushita, Y.: A closer look at
  rotation-invariant deep point cloud analysis. In: ICCV (2021)

\bibitem{li2018so}
Li, J., Chen, B.M., Lee, G.H.: So-net: Self-organizing network for point cloud
  analysis. In: CVPR (2018)

\bibitem{li2020end}
Li, L., Zhu, S., Fu, H., Tan, P., Tai, C.L.: End-to-end learning local
  multi-view descriptors for 3d point clouds. In: CVPR (2020)

\bibitem{li2021rotation}
Li, X., Li, R., Chen, G., Fu, C.W., Cohen-Or, D., Heng, P.A.: A
  rotation-invariant framework for deep point cloud analysis. IEEE Trans. on
  Visualization and Computer Graphics  (2021)

\bibitem{li2018pointcnn}
Li, Y., Bu, R., Sun, M., Wu, W., Di, X., Chen, B.: Pointcnn: Convolution on
  x-transformed points. In: NeurIPS (2018)

\bibitem{li2021contextual}
Li, Y., Yao, T., Pan, Y., Mei, T.: Contextual transformer networks for visual
  recognition. IEEE Trans. on PAMI  (2022)

\bibitem{lin2021learning}
Lin, Z.H., Huang, S.Y., Wang, Y.C.F.: Learning of 3d graph convolution networks
  for point cloud analysis. IEEE Trans. on PAMI  (2021)

\bibitem{liu2018deep}
Liu, M., Yao, F., Choi, C., Sinha, A., Ramani, K.: Deep learning 3d shapes
  using alt-az anisotropic 2-sphere convolution. In: ICLR (2018)

\bibitem{liu2019densepoint}
Liu, Y., Fan, B., Meng, G., Lu, J., Xiang, S., Pan, C.: Densepoint: Learning
  densely contextual representation for efficient point cloud processing. In:
  ICCV (2019)

\bibitem{liu2019relation}
Liu, Y., Fan, B., Xiang, S., Pan, C.: Relation-shape convolutional neural
  network for point cloud analysis. In: CVPR (2019)

\bibitem{liu2020closer}
Liu, Z., Hu, H., Cao, Y., Zhang, Z., Tong, X.: A closer look at local
  aggregation operators in point cloud analysis. In: ECCV (2020)

\bibitem{long2022stand}
Long, F., Qiu, Z., Pan, Y., Yao, T., Luo, J., Mei, T.: Stand-alone inter-frame
  attention in video models. In: CVPR (2022)

\bibitem{maturana2015voxnet}
Maturana, D., Scherer, S.: Voxnet: A 3d convolutional neural network for
  real-time object recognition. In: IROS (2015)

\bibitem{mo2019partnet}
Mo, K., Zhu, S., Chang, A.X., Yi, L., Tripathi, S., Guibas, L.J., Su, H.:
  Partnet: A large-scale benchmark for fine-grained and hierarchical part-level
  3d object understanding. In: CVPR (2019)

\bibitem{NEURIPS2019_9015}
Paszke, A., Gross, S., Massa, F., Lerer, A., Bradbury, J., Chanan, G., Killeen,
  T., Lin, Z., Gimelshein, N., Antiga, L., Desmaison, A., Kopf, A., Yang, E.,
  DeVito, Z., Raison, M., et~al.: Pytorch: An imperative style,
  high-performance deep learning library. In: NeurIPS (2019)

\bibitem{poulenard2019effective}
Poulenard, A., Rakotosaona, M.J., Ponty, Y., Ovsjanikov, M.: Effective
  rotation-invariant point cnn with spherical harmonics kernels. In: 3DV (2019)

\bibitem{qi2017pointnet}
Qi, C.R., Su, H., Mo, K., Guibas, L.J.: Pointnet: Deep learning on point sets
  for 3d classification and segmentation. In: CVPR (2017)

\bibitem{qi2017pointnet++}
Qi, C.R., Yi, L., Su, H., Guibas, L.J.: Pointnet++: Deep hierarchical feature
  learning on point sets in a metric space. In: NIPS (2017)

\bibitem{qiu2022mlp}
Qiu, Z., Yao, T., Ngo, C.W., Mei, T.: Mlp-3d: A mlp-like 3d architecture with
  grouped time mixing. In: CVPR (2022)

\bibitem{rao2019spherical}
Rao, Y., Lu, J., Zhou, J.: Spherical fractal convolutional neural networks for
  point cloud recognition. In: CVPR (2019)

\bibitem{shen20203d}
Shen, W., Zhang, B., Huang, S., Wei, Z., Zhang, Q.: 3d-rotation-equivariant
  quaternion neural networks. In: ECCV (2020)

\bibitem{shen2018mining}
Shen, Y., Feng, C., Yang, Y., Tian, D.: Mining point cloud local structures by
  kernel correlation and graph pooling. In: CVPR (2018)

\bibitem{shi2020pv}
Shi, S., Guo, C., Jiang, L., Wang, Z., Shi, J., Wang, X., Li, H.: Pv-rcnn:
  Point-voxel feature set abstraction for 3d object detection. In: CVPR (2020)

\bibitem{simonyan2014very}
Simonyan, K., Zisserman, A.: Very deep convolutional networks for large-scale
  image recognition. In: ICLR (2015)

\bibitem{su2018splatnet}
Su, H., Jampani, V., Sun, D., Maji, S., Kalogerakis, E., Yang, M.H., Kautz, J.:
  Splatnet: Sparse lattice networks for point cloud processing. In: CVPR (2018)

\bibitem{sun2019srinet}
Sun, X., Lian, Z., Xiao, J.: Srinet: Learning strictly rotation-invariant
  representations for point cloud classification and segmentation. In: ACM MM
  (2019)

\bibitem{tchapmi2017segcloud}
Tchapmi, L., Choy, C., Armeni, I., Gwak, J., Savarese, S.: Segcloud: Semantic
  segmentation of 3d point clouds. In: 3DV (2017)

\bibitem{thomas2019kpconv}
Thomas, H., Qi, C.R., Deschaud, J.E., Marcotegui, B., Goulette, F., Guibas,
  L.J.: Kpconv: Flexible and deformable convolution for point clouds. In: ICCV
  (2019)

\bibitem{touvron2021training}
Touvron, H., Cord, M., Douze, M., Massa, F., Sablayrolles, A., J{\'e}gou, H.:
  Training data-efficient image transformers \& distillation through attention.
  In: ICML (2021)

\bibitem{wang2019dynamic}
Wang, Y., Sun, Y., Liu, Z., Sarma, S.E., Bronstein, M.M., Solomon, J.M.:
  Dynamic graph cnn for learning on point clouds. Acm Trans. On Graphics
  \textbf{38}(5),  1--12 (2019)

\bibitem{weiler20183d}
Weiler, M., Geiger, M., Welling, M., Boomsma, W., Cohen, T.S.: 3d steerable
  cnns: Learning rotationally equivariant features in volumetric data. In:
  NeurIPS (2018)

\bibitem{wu2019pointconv}
Wu, W., Qi, Z., Fuxin, L.: Pointconv: Deep convolutional networks on 3d point
  clouds. In: CVPR (2019)

\bibitem{wu20153d}
Wu, Z., Song, S., Khosla, A., Yu, F., Zhang, L., Tang, X., Xiao, J.: 3d
  shapenets: A deep representation for volumetric shapes. In: CVPR (2015)

\bibitem{xiao2021triangle}
Xiao, C., Wachs, J.: Triangle-net: Towards robustness in point cloud learning.
  In: WACV (2021)

\bibitem{xu2021sgmnet}
Xu, J., Tang, X., Zhu, Y., Sun, J., Pu, S.: Sgmnet: Learning rotation-invariant
  point cloud representations via sorted gram matrix. In: ICCV (2021)

\bibitem{xu2021paconv}
Xu, M., Ding, R., Zhao, H., Qi, X.: Paconv: Position adaptive convolution with
  dynamic kernel assembling on point clouds. In: CVPR (2021)

\bibitem{xu2018spidercnn}
Xu, Y., Fan, T., Xu, M., Zeng, L., Qiao, Y.: Spidercnn: Deep learning on point
  sets with parameterized convolutional filters. In: ECCV (2018)

\bibitem{yang2019std}
Yang, Z., Sun, Y., Liu, S., Shen, X., Jia, J.: Std: Sparse-to-dense 3d object
  detector for point cloud. In: ICCV (2019)

\bibitem{you2018pvnet}
You, H., Feng, Y., Ji, R., Gao, Y.: Pvnet: A joint convolutional network of
  point cloud and multi-view for 3d shape recognition. In: ACM MM (2018)

\bibitem{yu2020deep}
Yu, R., Wei, X., Tombari, F., Sun, J.: Deep positional and relational feature
  learning for rotation-invariant point cloud analysis. In: ECCV (2020)

\bibitem{zhang2019rotation}
Zhang, Z., Hua, B.S., Rosen, D.W., Yeung, S.K.: Rotation invariant convolutions
  for 3d point clouds deep learning. In: 3DV (2019)

\bibitem{zhang2019shellnet}
Zhang, Z., Hua, B.S., Yeung, S.K.: Shellnet: Efficient point cloud
  convolutional neural networks using concentric shells statistics. In: ICCV
  (2019)

\bibitem{zhao2019pointweb}
Zhao, H., Jiang, L., Fu, C.W., Jia, J.: Pointweb: Enhancing local neighborhood
  features for point cloud processing. In: CVPR (2019)

\bibitem{zhao2021point}
Zhao, H., Jiang, L., Jia, J., Torr, P.H., Koltun, V.: Point transformer. In:
  ICCV (2021)

\bibitem{zhou2018voxelnet}
Zhou, Y., Tuzel, O.: Voxelnet: End-to-end learning for point cloud based 3d
  object detection. In: CVPR (2018)

\end{thebibliography}

\end{document}